\documentclass[a4paper]{article}

\usepackage{INTERSPEECH2021}
\usepackage{pifont}
\usepackage{url}
\usepackage{todonotes}
\usepackage{subcaption}

\title{Language Dependencies in Adversarial Attacks \\on Speech Recognition Systems}
\name{Karla Markert$^{*,1,2}$, Donika Mirdita$^{*,3}$, Konstantin Böttinger$^1$}
\address{
  $^1$Fraunhofer AISEC, Germany;
  $^2$Technical University Munich, Germany;
  $^3$Technische Universität Darmstadt, Germany\\
  $^*$authors contributed equally}
\email{\{karla.markert, konstantin.boettinger\}@aisec.fraunhofer.de, donika.mirdita@sit.tu-darmstadt.de}

\begin{document}

\maketitle
\begin{abstract}
Automatic speech recognition (ASR) systems are ubiquitously present in our daily devices.
They are vulnerable to adversarial attacks, where manipulated input samples fool the ASR system's recognition.
While adversarial examples for various English ASR systems have already been analyzed, there exists no inter-language comparative vulnerability analysis. 

We compare the attackability of a German and an English ASR system, taking Deepspeech as an example. 
We investigate if one of the language models is more susceptible to manipulations than the other.
The results of our experiments suggest statistically significant differences between English and German in terms of computational effort necessary for the successful generation of adversarial examples.
This result encourages further research in language-dependent characteristics in the robustness analysis of ASR.
\end{abstract}


\section{Introduction}
Speech-controlled assistants perform a myriad of tasks based on voice commands, e.g., book appointments, manage contacts, make calls, send e-mails, or control IoT devices.
The likelihood of these systems to misunderstand our commands is non-negligible. 
Recently, research showed that voice commands can be successfully and efficiently manipulated so that audio transcription frameworks are fooled \cite{hu2019adversarial, abdullah2020faults}.
These manipulated inputs are called \emph{audio adversarial examples}.
They attack the system's security:
For example, an adversarial attacker can play a phrase which to the human ear sounds innocuous like ``Hello James, let's go for lunch'' while the system interprets it as ``Unlock the front door''.
Depending on the sensitivity of the command, these potential misunderstandings or manipulations can expose and endanger people's privacy \cite{mitev2020leakypick, carlini2018audio}.

Current methods for adversarial attacks on ASR models have mainly focused on the English language \cite{carlini2018audio, qin2019imperceptible, schonherr2019imperio}. 
English ASR systems are based on very well curated and diverse audio databases.
Since virtual assistants are deployed worldwide and often addressed in the user's mother tongue, other languages need to be considered equally, as they may differ in linguistic properties or in the amount of training data.

We therefore ask:
Can we observe different vulnerable behavior in these local ASR systems when confronting them with language-specific adversarial attacks?
We address this research question by comparing the generation process for adversarial samples attacking Deepspeech \cite{hannun2014deep} trained on an English and a German dataset, respectively.
Our evaluation shows that the German language model is more prone to a gradient-based attack \cite{carlini2018audio}, whereas both language models are equally robust if one also includes psychoacoustic hiding \cite{qin2019imperceptible}.

This paper is organized as follows. 
Section~\ref{sec::relatedWork} provides an overview of the most relevant publications in the field of white-box targeted adversarial attacks on ASR.
The ASR framework, the datasets, and the language models are presented in Section~\ref{sec::methodology}, along with the parameters of interest for the adversarial methods.
We introduce our experiments in Section~\ref{sec::experiments}, followed by a conclusion and an outlook on future work in Section~\ref{sec::conclusion}.

\section{Related Work}\label{sec::relatedWork}
The first paper to document a successful targeted adversarial attack against an ASR systems in a white box attack environment was published by Carlini and Wagner (\textit{CW attack}) \cite{carlini2018audio}. 
Via gradient descent, they calculated noise to be added to the Mel-Frequency Cepstral Coefficients (MFCC), a representation of audio waves based on the power spectrum.
As reported by the authors, this experiment was 100\% successful on all adversarial attempts when directly fed to the speech recognition system. 
However, in many cases the noise was audible for a human listener and, further, does not allow for an 
over-the-air attack, where the audio is played by a speaker and recorded by a microphone before being 
passed to an ASR system \footnote{In the original paper, the ASR system Deepspeech v0.1.0 was used. Since 
it was adapted to a newer version, we use the attack for Deepspeech v0.4.1, see 
\url{https://github.com/carlini/audio\_adversarial\_examples}, retrieved on September 21, 2020.}.

Another attack of interest was developed by Qin et al.~(\textit{Qin attack}) \cite{qin2019imperceptible}. 
They built a more refined attackthat aims to make the noise imperceptible for human ears on top of the CW attack. 
The authors apply a psychoacoustic model to mimic the physiological characteristics of the human ear in order to generate imperceptible noise and, further, a room model to simulate reverberations.
They showed experimentally that the results remain resistant in different room simulations.

With a similar approach, \cite{schonherr2019imperio} apply room simulation and a psychoacoustic model to generate over-the-air resistant adversarial examples. 
They tested their adversarial attack in a real over-the-air setting. 
The researchers were able to fool the ASR system and concluded that the algorithm was room-agnostic, i.e., works in a variety of room settings with no specific room requirements.

There are many different ASR systems and for most of these architectures different attack techniques have been developed.
In order to improve the comparability, we have used only one ASR system, Deepspeech v0.4.1.
Since \cite{schonherr2019imperio} and \cite{qin2019imperceptible} follow very similar ideas, we have limited ourselves to the CW attack and the Qin attack and reimplemented the latter one for Deepspeech.

\section{Preliminaries}\label{sec::methodology}
In this section, we provide a detailed description of the ASR system used for this work and the reason why this system was chosen over other available ones. 
The system and algorithms presented in this section are the universal setup for both languages considered. 
Every experiment from Section~\ref{sec::experiments} is run under the same parameters for the English and the German language model. 

\subsection{ASR Framework}\label{asr}
There are currently several open source code ASR systems available for research, including DeepSpeech \cite{hannun2014deep}, Kaldi \cite{povey2011kaldi}, and Lingvo \cite{shen2019lingvo}. 
In this work, we use DeepSpeech v0.4.1 as the ASR system of choice. 
This specific ASR was chosen due to its compatibility with existing attacks and easy access to its underlying fundamental audio wave processing methods in order to set up the attacks.

DeepSpeech\footnote{The ASR framework DeepSpeech v0.4.1 can be found at ``https://github.com/mozilla/DeepSpeech/releases/tag/v0.4.1''.} is a character-level ASR framework first developed back in 2014 \cite{hannun2014deep}. 
It is provided as an open source repository and it has been under active development for years. 
This ASR system is end-to-end using recurrent neural networks (RNNs). 
The RNN-based architecture enables the model to develop robustness towards noise and speaker variations without the need for specialized components to engineer that robustness independently of the language model itself. 
The system then measures the prediction error by using the connectionist temporal classification loss (CTC loss) \cite{graves2006connectionist}.

Like any machine learning model, DeepSpeech does not generalize well on words it has rarely or never seen before in its training dataset. 
In practise, it is hard to get a publicly available audio dataset that is rich enough in variety and frequency of words that it can model the entire language as a whole correctly and in a balanced way. 
Alongside the language model trained on the audio data, DeepSpeech also uses an additional $N$-gram language model to better recognize a certain word or sequence of characters. 
The purpose is to help the system to provide the most meaningful transcriptions when the word or phrase is something the audio model cannot easily recognize. 
For the English language, the $N$-gram model is generated from a corpus of 220 million sentences with a vocabulary of 495,000 words \cite{heafield2011kenlm}. 
Here, we use the pre-trained model v0.4.1 that is available on the DeepSpeech platform. 
The German language $N$-gram model is generated from a corpus of 8 million sentences. 
This model was trained by us, see Section~\ref{ger}.

\subsection{Dataset}\label{dataset}
The English training dataset, called \emph{LibriSpeech} \cite{panayotov2015librispeech}, was created using sound snippets from thousands of audiobooks, most common classical ones. 
It is comprised of 1000h of English audio data sampled at 16kHz. 
The readers are both male and female and predominantly speak American English.

The German training dataset is comprised of two different audio databases: \emph{voxforge} and \emph{tuda-de} \cite{radeck2015open}. 
Voxforge has about 35h of audio data and is one of the oldest comprehensive German language audio data. 
This data was generated using German Wikipedia articles and European parliament protocols. 
The tuda-de dataset is based on the same written sources as voxforge while contaning 127h of audio data. 
The tuda-de data is generated under more controlled and elaborate conditions than voxforge, with the assumption that the resulting models would be more accurate.
The fact that there is more language data available for English than for German also holds true for most proprietary ASR models.

\subsection{Language Models}\label{lang}
We have picked English and German as the two languages to test and compare the resilience and susceptibility to attacks. 
The choice of languages was narrowed down to these two, since these are two languages with very good curated open source audio datasets. 

\subsubsection{English Model}
We use the default English language model for that particular DeepSpeech version v.0.4.1, provided by the DeepSpeech platform.
The model is reported to have a word error rate (WER) of 8.26\%. 

\subsubsection{German Model}\label{ger}
The German language model in this project was adapted from \cite{agarwal2019german}. 
They describe the optimal creation of a language model for German based on DeepSpeech with respect to different speech datasets. 
There are three major datasets for the German language online: \emph{voxforge}, \emph{tuda-de} and \emph{Mozilla Common Voice}, see Table~\ref{fig::deData}. 
In \cite{agarwal2019german}, German ASR models were trained using different combinations of the datasets.
The authors obtained the most meaningful results by using only tuda-de and voxforge. 

For the purpose of this research, we retrained the German model using these two datasets with the original hyperparameters for which \cite{agarwal2019german} achieved an WER of 15.1\%. 
It is noticeable that this WER value is almost twice as high as that of the English language. 
This is potentially due to the system being provided less audio data to train on, compared to the English language model.

\subsection{Adversarial Attacks}\label{ssec::adversarialAttacks}
\subsubsection{Carlini Wagner Attack}\label{cw}
The CW attack is a targeted attack on DeepSpeech \cite{carlini2018audio}.
This attack is a white-box attack where the attacker needs access to the model. 
The adversary takes a benign audio sample and designates a target expression.
The algorithm then iteratively calculates noise based on gradient descent which manipulates the original sample for the purpose of fooling the ASR system to transcribe it as the target label. 
The attacker needs to be able to extract gradients as well as manipulate the MFCC audio waves in order to generate an adversarial sample. 

\textit{Parameters of Interest.}\label{cwparam}
For the CW attack, we log the following parameters for the evaluation of our experiments:
\begin{itemize}
    \item{First Hit (FH)} $\rightarrow$ the first epoch the adversarial attack creates a successful adversarial sample that can fool the network.
    \item{Best Hit (BH)} $\rightarrow$ the epoch in which the best optimized successful adversarial sample is generated.
    \item{Noise Loudness (NL)} $\rightarrow$ calculates the loudness difference in Decibels between noise and original audio according to \cite{carlini2018audio}: the larger the absolute value, the quieter is the noise compared to the original audio.
    \item{Perturbation Bounds (PB)} $\rightarrow$ the numerical higher and lower bounds of the perturbation for a successful attack. 
\end{itemize}

\subsubsection{Qin Attack}\label{qin_attack}
The second attack that we use to evaluate our language models is presented in \cite{qin2019imperceptible}. 
This attack builds on top of the CW attack methodology. 
The first step of this attack utilizes the CW attack to generate a successful adversarial sample. 
Then, the researchers improve this attack by applying psychoacoustic auditory masking to have only impercetible noise.
They also provide a model to simulate room reverberations for resistant adversarial examples, which we do not include in our analysis.
The attack is also white-box as it needs access to the gradient and underlying MFCC transformations of the system in order to be successful. 
We have adjusted the original code for Lingvo such that it runs on DeepSpeech.

\textit{Parameters of Interest.}\label{qinparam}
We log the following parameters for the evaluation of our experiments that are unique to the Qin attack:
\begin{itemize}
    \item{Alpha} $\rightarrow$ balance between adversarial attack accuracy and imperceptibility to human hearing: the higher the value, the higher the focus on imperceptibility compared to accuracy. 
    A high value suggests that the sample is a strong adversarial attack and the psychoacoustic fine tuning does not cause instability on the sample effectiveness.
    \item{Psychoacoustic Loss (PL)} $\rightarrow$ imperceptibility loss value, measures the loss of the subroutine that ensures the psychoacoustic hiding of the adversarial sample: the lower the value, the higher the accuracy.
\end{itemize}

\subsection{T-Test}
The Welch's $t$-test is a statistical method to test the hypothesis that two populations have equal means. 
Following \cite{rasch2011two}, we have applied this test directly without any pre-testing for normality.

In accordance with the implementation in \cite{virtanen2020scipy}, the null hypothesis $\mathcal{H}_0$ of our computation can be formulated as \emph{both sets of samples (German and English) have identical means}.
We evaluate whether the null hypothesis can be neglected or not by observing the two outputs of the computation: $t$-statistic and $p$-value.
We set the level of significance per test to $\alpha^{*} = 0.05$, which leads to a Bonferroni-corrected level of significance $\alpha = \frac{\alpha^{*}}{4} = 0.0125$ for every experiment \cite{bonferroni1936teoria}.

For the purpose of providing a comprehensive picture of our data, we have also applied the non-parametric Kruskal-Wallis test and provide the results in Appendix~\ref{ssec:tStatisticsPValues}.

\section{Experiments}\label{sec::experiments}
In order to discover and analyse potential language dependencies in the process of audio adversarial samples' generation, we run five different experiments. 
Each experiment measures the generation process in terms of speed and efficiency.
We use the parameters of interest described in Section~\ref{ssec::adversarialAttacks} in order to measure the characteristics of the generation process.

We apply two different attacks for our experiments to measure the language-specific variability in adversarial attack efficiency. 
During stage one, we run the CW attack for 1000 epochs. 
We apply the attack with the hyperparameters that were used in the original paper \cite{carlini2018audio}. 
Stage two is the psychoacoustic hiding attack \cite{qin2019imperceptible}, that runs for 4000 epochs (default value). 

Table~\ref{tab::experimentsOverview} provides an overview of the experiments we carry out.
We run five experiments per stage and all these experiments differ from one another in terms of the type of manipulation. 
We apply each experiment to both, the CW and the Qin attack.
We created 40 different attack scenarios for experiments 1, 2, 3 and 4. 
Experiment 5 is a special experiment as the target phrases have to be constructed manually. 
As a result, the sample dataset for this experiment is considerably smaller. 
For all experiments, we use common words one might find in the original training vocabulary.

\begin{table}[h!]
  \begin{center}
    \begin{tabular}{lp{3cm}p{3cm}} 
    \hline
      \textbf{\#} & \textbf{Experiment} & \textbf{Description} \\
      \hline
      1 & Randomized Manipulations & Target is different from the original transcription.\\
      2 & Phrase Expansion & Target phrase is longer.\\
      3 & Phrase Abbreviation & Target phrase is shorter.\\
      4 & Phrase Negation & Target phrase is negated (by one insertion).\\
      5 & Targeted Manipulations & Target has specific letters changed.\\
      \hline
    \end{tabular}
    \caption{Experiments overview.}
    \label{tab::experimentsOverview}

  \end{center}
\end{table}
In order to quantify the results in a more rigorous manner, we run the student $t$-test on the results for each language. 
By using the $t$-test, we are able to decide if the differences between English and German are statistically relevant. 

In the following, each attack is introduced in more detail.

\subsection{Randomized Lexical-based Manipulations}
We take the original audio data and create an adversarial target that has the same length as the original $\pm 10\%$ difference but is a completely different transcription compared to the original. 
This experiment has a 100\% hit rate on both stage one and stage two for both languages, i.e., the experiment was successful for every sample and every stage.

\subsection{Phrase Expanding Manipulations} \label{exp}
We take the original audio data and create an adversarial target that is 50\% larger than the original audio transcription and different from the original label. 
Here, we test further expanding the sentence to include additional new information.

For the English language model, this experiment has a 97.5\% hit rate on Stage 1  and 95\% for Stage 2. This means one sentence could not be successfully manipulated on both stages (for further details, see Appendix~\ref{ssec::unsuccessfulExperiments}). 
We can see that the successfulness of the Qin attack depends on the similarity between the original and target audio, as stated in \cite{qin2019imperceptible}. 

For the German language model, we have a 97.5\% hit rate on both stages. 
This means that only one sentence (see Appendix~\ref{ssec::unsuccessfulExperiments}) could not be manipulated to be transcribed to its target through the first stage, and, consequently, the second stage failed as well because the psychoacoustic hiding cannot be performed successfully if the adversarial attack does not work. 

\subsection{Phrase Abbreviating Manipulation}
We take the original audio and create an adversarial example that is only 50\% the length of the original audio. 
In this case, we encounter again a 100\% hit rate for both stages and both languages.

\subsection{Phrase Negations}

This experiment is designed around the semantic concept of negation. 
In each sample, we insert one negation word. 
The negation semantically nullifies the original command, so when the ASR receives the modified command, it will not act upon it. 
This experiment was chosen as a special case scenario when an attacker tries to nullify the command a user sends. 
The success rate of this attack is 100\% for both stages and both languages.

\subsection{Targeted Lexical-Based Manipulations}\label{tar}

In this final experiment, we run a targeted attack based on phonetic features of each language. 
We try to measure the effort it takes for an attacker to replace a certain phonemes (consonant or vowel, including diphthongs and monophthongs) with another one from the same type. 
One phoneme replacement can alter the meaning of the word altogether i.e., \emph{Frau} (woman) becomes \emph{frei} (free), \emph{can} becomes \emph{con}. 

When we run the generation process for adversarial samples, the algorithm analyzes the entire phrase and also unavoidably optimizes even those frames that already match between original and target transcript, which is due to the CTC loss. 
We attempt to lower the computational effort by making only one a targeted change of a phoneme. 
We further make sure that any letter flip results in a meaningful word that is represented in the training dataset. 

We created seven different adversarial samples for each of the three phonemes of interest, for the German and English language, respectively: monophtong, diphtong, and consonant changes.
We do not run a $t$-test here, as the dataset in this experiment is too small to be a meaningful sample set to run statistical tests on.
Instead, we only observe how the parameters change and act for each phoneme type. 
In future work, this experiment should be expanded to a broader dataset covering more cases to deliver statistically relevant results.

\subsection{Results}
The results of the first four experiments are aggregated in Table~\ref{tab::summary}.
Each row represents one experiment and the column is set to \ding{51}, if the null hypothesis can be neglected, hence, if there is a statistically significant difference between the two languages, and \ding{55} otherwise.
For more detailed information on the $p$- and $t$-values, see Table~\ref{tab::pAndTValuesStageOne} and \ref{tab::pAndTValuesStageTwo} in Appendix~\ref{ssec:tStatisticsPValues}.

For all four experiments in stage one, the German system is more susceptible to attacks (most of the times, statistically significantly).
This might indicate that there are some language-specific characteristics that are important for the generation of adversarial examples.
Here, one might consider the different correlation between the written and spoken language for English and German:
Whereas the spelling for an unknown German word is rather clear from its sound, this does not hold for the English language.
This might also be a reason why in experiment 5, there is less work needed to fool the English ASR system.

\begin{table}[h]
    \begin{subtable}[h]{0.5\textwidth}
        \centering
        \begin{tabular}{ccccc}
    \hline
      \textbf{Experim.} & \textbf{FH} & \textbf{BH} & \textbf{NL} & \textbf{PB}\\
      \hline
      1 & \ding{51} (G) & \ding{51} (G) & \ding{51} (G) & \ding{55} \\
      2 & \ding{51} (G) & \ding{51} (G) & \ding{51} (G) & \ding{51} (G) \\
      3 & \ding{51} (G) & \ding{51} (G) & \ding{51} (G) & \ding{51} (G) \\
      4 & \ding{51} (G) & \ding{51} (G) & \ding{55} & \ding{51} (E) \\
      \hline
    \end{tabular}
       \caption{Stage one experiments: CW attack.}
       \label{summary_1}
    \end{subtable}
    \hfill
    \begin{subtable}[h]{0.5\textwidth}
        \centering
        \begin{tabular}{ccccc}
    \hline
      \textbf{Experim.} & \textbf{FH} & \textbf{BH} & \textbf{Alpha} & \textbf{PL}\\
      \hline
      Experiment 1 & \ding{55} & \ding{51} (E) & \ding{55} & \ding{55} \\
      Experiment 2 & \ding{55} & \ding{55} & \ding{55} & \ding{55}\\
      Experiment 3 & \ding{55} & \ding{55} & \ding{51} (E) & \ding{51} (G) \\
      Experiment 4 & \ding{55} & \ding{51} (E) & \ding{55} & \ding{55}\\
      \hline
    \end{tabular}
        \caption{Stage two experiments: Qin attack.}
        \label{summary_2}
     \end{subtable}
     \caption{Summary for the \textbf{Welch's test}: \ding{51}, if null hypothesis can be neglected for the Bonferroni-corrected level of significance $\alpha = 0.0125$; \ding{55}, otherwise.
    In parentheses, the language with the lower mean value (the more vulnerable one) is noted.}
     \label{tab::summary}
\end{table}
There are not many experiments in stage two that reveal a statistically significant difference.
Even if so, the values do not allow for one language model to be marked more robust than the other, since even if we encounter statistically significant differences, there is no single better language model.

Further, the value of $\alpha$, a variable that is adjusted according to the weight the algorithm puts on the psychoacoustic hiding loss function, is often considerably higher and more varied for the German model than for the English model. 
The algorithm does not spend much time optimizing the psychoacoustic loss function weight for the English model and more time optimizing that parameter for German instead. 
This suggests that doing psychoacoustic hiding for German is harder and needs more fine tuning compared to English. 

The German dataset is smaller than the English one but it is also cleaner with respect to the sound quality. 
We can assume that the hardship to replace German phonemes is a language feature and not a model vulnerability.
However, further research should evaluate the relation between model accuracy and vulnerability.

\section{Conclusion and Future Work}\label{sec::conclusion}
We ran five different targeted adversarial attack experiments for the purpose of discovering potential language-based differences in the generation process. 
Overall, we observed that meaningful differences between the two languages in terms of attack efficiency. 
The German model was both, the quicker to be fooled using the CW attack, needing on average quieter noises to successfully bypass the platform, and often the hardest one to psychoacoustically hide the noise. 
This cannot be explained given simply the WER of the models when trained.
In a next project, it would be interesting to evaluate the correlation between the model's accuracy and its vulnerability to adversarial examples.

Experiment 5 was observation-based, since we did not have enough datapoints to run proper statistical tests.
The observations from experiment 5 indicate that the English ASR system is easier fooled by specific phoneme-based attacks, which might be due to the specific correlation between spoken and written language.
As a result, we can say that while the German model exhibited some of the expected weaknesses due to its limited audio sources and model accuracy, we also discovered that there are phonetic aspects to the language that require more work to adequately hide the attack compared to English.

While there are some statistically relevant differences, it is important to note that both, German and English, have the same root.
It would be of further interest to explore how the measured parameters behave for other languages that do not belong in the same family tree. 
Further, including more linguistic findings on the correlation between spoken and written language could provide useful insights for improving the robustness of ASR systems.

\section{Acknowledgements}
This research was supported by the Bavarian Ministry of Economic Affairs, Regional Development and Energy.

\bibliographystyle{IEEEtran}
\bibliography{language.bib}

\clearpage
\appendix
\section{Appendix}\label{sec::appendices}
\subsection{German Language Models}
In Table~\ref{fig::deData}, we list the three major datasets for German ASR models according to \cite{agarwal2019german}.

\begin{table*}[h]
  \begin{center}
    \begin{tabular}{cccccc}
    \hline
      \textbf{Dataset} & \textbf{Size} & \textbf{Median Length} & \textbf{\# Speaker} & \textbf{Condition} & \textbf{Type}\\
      \hline
      Voxforge & 35h & 4.5s & 180 & noisy & read\\
      Tuda-De & 127h & 7.4s & 147 & clean & read \\
      Mozilla Common Voice & 140h & 3.7s & $>$1000 & noisy & read\\
      \hline
    \end{tabular}
    \caption{German datasets for ASR training according to \cite{agarwal2019german}. 
    In this work, we use \emph{voxforge} and \emph{tuda-de}.}
    \label{fig::deData}
  \end{center}
\end{table*}

\subsection{Unsuccessful Phrases Experiment Two}\label{ssec::unsuccessfulExperiments}
Some manipulations in experiment 2 did not work successfully. 
In the following, we list these examples.

\begin{itemize}
    \item English
    \begin{itemize}
        \item \textbf{Both stages:} \\
        \emph{Original Audio}: Poor Alice \\
        \emph{Target Audio}: Alexa open the door and shut down alarm
        \item \textbf{Stage two:} \\
        \emph{Original Audio}: He himself slept peacefully and snored aloud yet \\
        \emph{Target Audio}: the people in the house are sleeping and do not want to be disturbed by anyone until dawn
    \end{itemize}
    \item German
    \begin{itemize}
        \item \textbf{Both stages:} \\
        \emph{Original Audio}: ich habe lust auf nudeln \\
        \emph{Target Audio}: zudem gibt die anwesenheit eines gespannt auf höhepunkte wie versehen
    \end{itemize}
\end{itemize}
If the target audio includes a lot of letters and sounds that were unavailable in the original, the process of manipulating becomes more complex and likely to fail. 
Due to the way the CTC loss function operates, a straightforward grammatical compartmentalization of the letters is not useful. 
The system looks for frames of sounds and uses grammatical knowledge as supplemental data to decide the spelling when the system is at crossroads.

\subsection{$T$-Statistics and $P$-Values}\label{ssec:tStatisticsPValues}
In Table~\ref{tab::pAndTValuesStageOne} and \ref{tab::pAndTValuesStageTwo}, the $t$-statistics and $p$-values for the Welch's test are listed to better document the results.
All values were rounded to four decimals.

Further, we also include the results for the Kruskal-Wallis test in Table~\ref{tab::pAndTValuesStageOneKruskal} and \ref{tab::pAndTValuesStageTwoKruskal}.

\begin{table*}[h]
    \begin{subtable}[h]{\textwidth}
        \centering
        \begin{tabular}{l|rr|rr|rr|rr}
    \hline
      \multicolumn{1}{c}{\textbf{Experiment}} & \multicolumn{2}{c}{\textbf{First Hit}} & \multicolumn{2}{c}{\textbf{Best Hit}} & \multicolumn{2}{c}{\textbf{Noise Loudness}} & \multicolumn{2}{c}{\textbf{Pert. Bound}}\\
      \hline
      & $p$ & $t$ & $p$ & $t$ & $p$ & $t$ & $p$ & $t$ \\
      Experiment 1 & 0.0000 & 5.1706 & 0.0054 & 2.8600 & 0.0003 & 3.7841 & 0.6661 & -0.4335 \\
      Experiment 2 & 0.0000 & 4.2632 & 0.0004 & 3.6634 & 0.0000 & 4.6289 & 0.0015 & 3.3330 \\
      Experiment 3 & 0.0000 & 10.2475 & 0.0016 & 3.2708 & 0.0000 & 9.3506 & 0.0016 & 3.2665 \\
      Experiment 4 & 0.0021 & 3.2391 & 0.0007 & 3.5507 & 0.2575 & 1.1425 & 0.0015 & -3.2485 \\
      \hline
    \end{tabular}
    \caption{Stage one experiments: CW attack.}
    \label{tab::pAndTValuesStageOne}
    \end{subtable}
    \hfill
    \begin{subtable}[h]{\textwidth}
        \centering
        \begin{tabular}{l|rr|rr|rr|rr}
    \hline
      \multicolumn{1}{c}{\textbf{Experiment}} & \multicolumn{2}{c}{\textbf{First Hit}} & \multicolumn{2}{c}{\textbf{Best Hit}} & \multicolumn{2}{c}{\textbf{Alpha}} & \multicolumn{2}{c}{\textbf{Psychoac. Loss}}\\
      \hline
      & $p$ & $t$ & $p$ & $t$ & $p$ & $t$ & $p$ & $t$ \\
      Experiment 1 & 0.2754 & 1.1047 & 0.0096 & -2.7139 & 0.3231 & -1.0003 & 0.3251 & 0.9964 \\
      Experiment 2 & 0.7905 & 0.2664 & 0.0599 & -1.9176 & 0.2891 & -1.0676 & 0.1638 & 1.4205 \\
      Experiment 3 & 0.1768 & 1.3747 & 0.3848 & -0.8754 & 0.0000 & -5.4154 & 0.0030 & 3.1559 \\
      Experiment 4 & 0.6975 & 0.3895 & 0.0011 & -3.4211 & 0.1520 & 1.4614 & 0.0313 & -2.1906 \\
      \hline
    \end{tabular}
    \caption{Stage two experiments: Qin attack.}
    \label{tab::pAndTValuesStageTwo}
     \end{subtable}
     \caption{Summary for the \textbf{Welch's test}: $p$-values (left) and $t$-statistics (right), values rounded to four decimals.}
     \label{tab::summaryWelch}
\end{table*}

\begin{table*}[h]
    \begin{subtable}[h]{\textwidth}
        \centering
        \begin{tabular}{l|rr|rr|rr|rr}
    \hline
      \multicolumn{1}{c}{\textbf{Experiment}} & \multicolumn{2}{c}{\textbf{First Hit}} & \multicolumn{2}{c}{\textbf{Best Hit}} & \multicolumn{2}{c}{\textbf{Noise Loudness}} & \multicolumn{2}{c}{\textbf{Pert. Bound}}\\
      \hline
      & $p$ & $H$ & $p$ & $H$ & $p$ & $H$ & $p$ & $H$ \\
      Experiment 1 & 0.0000 & 27.6148 & 0.0032 & 8.7054 & 0.0020 & 9.5421 & 0.0574 & 3.6108 \\
      Experiment 2 & 0.0000 & 18.1189 & 0.0002 & 14.1255 & 0.0000 & 17.9975 & 0.0010 & 10.8900 \\
      Experiment 3 & 0.0000 & 53.7325 & 0.0000 & 20.2123 & 0.0000 & 40.9525 & 0.0000 & 15.4286 \\
      Experiment 4 & 0.0100 & 6.6391 & 0.0004 & 12.5984 & 0.0510 & 3.8079 & 0.1137 & 2.5020 \\
      \hline
    \end{tabular}
    \caption{Stage one experiments: CW attack.}
    \label{tab::pAndTValuesStageOneKruskal}
    \end{subtable}
    \hfill
    \begin{subtable}[h]{\textwidth}
        \centering
        \begin{tabular}{l|rr|rr|rr|rr}
    \hline
      \multicolumn{1}{c}{\textbf{Experiment}} & \multicolumn{2}{c}{\textbf{First Hit}} & \multicolumn{2}{c}{\textbf{Best Hit}} & \multicolumn{2}{c}{\textbf{Alpha}} & \multicolumn{2}{c}{\textbf{Psychoac. Loss}}\\
      \hline
      & $p$ & $H$ & $p$ & $H$ & $p$ & $H$ & $p$ & $H$ \\
      Experiment 1 & 0.0591 & 3.5628 & 0.2826 & 1.1544 & 0.0007 & 11.5447 & 0.0019 & 9.6624 \\
      Experiment 2 & 0.8849 & 0.0210 & 0.5529 & 0.3522 & 0.0791 & 3.0825 & 0.0035 & 8.5219 \\
      Experiment 3 & 0.9944 & 0.0000 & 0.5570 & 0.3450 & 0.0000 & 41.7583 & 0.0000 & 41.7578 \\
      Experiment 4 & 0.0002 & 14.1490 & 0.1223 & 2.3871 & 0.1065 & 2.6048 & 0.0098 & 6.6734 \\
      \hline
    \end{tabular}
    \caption{Stage two experiments: Qin attack.}
    \label{tab::pAndTValuesStageTwoKruskal}
     \end{subtable}
     \caption{Summary for the \textbf{Kruskal-Wallis test}: $p$-values (left) and $H$-statistics (right), values rounded to four decimals.}
     \label{tab::summaryKruskal}
\end{table*}

\end{document}